\documentclass{article} 
\usepackage{colm2024_conference}
\usepackage{microtype}
\usepackage{hyperref}
\usepackage{url}
\usepackage{booktabs}
\usepackage[most]{tcolorbox}

\colmfinalcopy 
\usepackage{inconsolata}
\usepackage{hyperref}       
\usepackage{url}            
\usepackage{booktabs}       
\usepackage{amsfonts}       
\usepackage{nicefrac}       
\usepackage{microtype}      

\usepackage{graphicx}
\usepackage{soul,color}
\usepackage{tabularx}
\usepackage{colortbl}
\usepackage{bm}

\usepackage{array}
\usepackage{listings}
\usepackage{makecell}
\usepackage{amsmath}
\usepackage{soul}

\definecolor{deepred}{rgb}{0.8, 0, 0}
\definecolor{shallowred}{rgb}{1, 0.8, 0.8}
\definecolor{greenbox}{rgb}{0.8, 1, 0.8}
\sethlcolor{pink}
\setstcolor{green}
\setulcolor{red}

\usepackage{listings}
\usepackage{array}

\lstnewenvironment{code}{\lstset{basicstyle=\ttfamily,frame=single}}{}

\newcommand{\fixshu}[1]{\footnote{\textcolor{red}{\textbf{FIX-Shu!!!} #1}}}
\newcommand{\fixme}[1]{\footnote{\textbf{FIXME!!!} #1}}
\newcommand{\OurMODEL}{\textsc{Monal}}

\newcommand{\eat}[1]{}

\newcommand{\warn}[1]{\textcolor{red}{#1}}
\DeclareUnicodeCharacter{0301}{\hspace{-1ex}\'{ }}

\title{\OurMODEL{}: Model Autophagy Analysis for Modeling Human-AI Interactions}

\author{
Shu Yang\thanks{The first two authors contributed equally to this work.}$^{*,1,2,3}$,
Muhammad Asif Ali$^{*,1,2}$,
Lu Yu$^{5}$,
Lijie Hu\thanks{Corresponding author.}$^{\dagger,1,2,4}$,
and Di Wang$^{\dagger,1,2,4}$ \\
$^1$Provable Responsible AI and Data Analytics (PRADA) Lab\\
$^2$King Abdullah University of Science and Technology\\
$^3$University of Macau \quad $^4$SDAIA-KAUST AI \quad $^5$Ant Group 
}

\begin{document}
\maketitle

\begin{abstract}
The increasing significance of large models and their multi-modal variants in societal information processing has ignited debates on social safety and ethics. However, there exists a paucity of comprehensive analysis for:
(i) the interactions between human and artificial intelligence systems, and (ii) understanding and addressing the associated limitations. To bridge this gap, we propose
\textbf{\underline{M}}odel 
\textsc{A}ut\textbf{\underline{O}}phagy \textsc{a}\textbf{\underline{NAL}}ysis (\OurMODEL{}) 
for large models' self-consumption explanation.
\OurMODEL{} employs two distinct autophagous loops (referred to as ``self-consumption loops'') to elucidate the suppression of human-generated information in the exchange between human and AI systems. Through comprehensive experiments on diverse datasets, we evaluate the capacities of generated models as both creators and disseminators of information.
Our key findings reveal (i) A progressive prevalence of model-generated synthetic information over time within training datasets compared to human-generated information; (ii) The discernible tendency of large models, when acting as information transmitters across multiple iterations, to selectively modify or prioritize specific contents; and (iii) The potential for a reduction in the diversity of socially or human-generated information, leading to bottlenecks in the performance enhancement of large models and confining them to local optima.
\end{abstract}

\eat{
This study investigates the biases and preferences for 
using humans and large models as the key links in the communication.}

\eat{The increasing significance of large language and multi-modal models in 
societal information processing has ignited debates on social safety and ethics. 
However, few studies have approached the analysis of these limitations from the comprehensive perspective of human and artificial intelligence system interactions. 
This study investigates biases and preferences when humans and large models 
are used as key links in communication. 
To achieve this, we design a multi-modal dataset and three different experiments 
to evaluate generative models in their roles as producers and disseminators 
of information. 
Our main findings highlight that synthesized information is more likely to be 
incorporated into model training datasets and messaging than human-generated 
information. 
Additionally, large models, when acting as transmitters of information, tend 
to modify and lose specific content selectively. 
Conceptually, we present two realistic models of autophagic ("self-consumption") 
loops to account for the suppression of human-generated information in the 
exchange of information between humans and AI systems. 
We generalize the declining diversity of social information and the 
bottleneck in model performance caused by the above trends to the 
local optima of large models.}

\section{Introduction}
Large models, including large language models (LLMs)~\citep{bai2022constitutional, zeng2022glm, openai2023gpt4, touvron2023llama} and large multi-modal models~\citep{yang2023dawn,yin2023survey}, are rapidly emerging as transformative tools, reshaping our world in a formidable way. Among their myriad implications, their growing social impact stands out, making them an integral component of our modern communication era. These models facilitate the dissemination of viewpoints and information within human society by engaging in continual interaction with humans~\citep{gao2023s3,bian2023influence}.
Particularly noteworthy are recent technological advancements that have sparked an arms race, resulting in the daily training of hundreds of next-generation models using a blend of real (human-generated) and synthetic (LLM-generated) data.
This iterative training process engenders an autophagous loop 
(elucidated in Section~\ref{sec:autop-def}) within the datasets, wherein new models are continually trained on synthetic data. Previous investigations by~\citet{alemohammad2023selfconsuming}, and~\citet{shumailov2023curse} 
reported the decline in data quality and diversity with repeatedly generated data, often employed to train visual generative models, a phenomenon termed Model 
Autophagy Disorder. They underscored the dearth of fresh and realistic training data as a primary driver of this disorder. However, their analyses entirely relied on simulated experiments to demonstrate the decline in model performance. Therefore, a deeper examination is warranted to elucidate why real data is becoming increasingly scarce and its implications for the flow of information in human society. We argue that despite their status as novel and consequential components of the communication era~\citep{edwards2016communication}, the inherent limitations of large models 
remain inadequately explored. Specifically, we aim to answer the following questions: (i) What impacts do human-generated real and synthetic data have on model training? (ii) To what extent do samples from repeatedly generated synthetic data influence data quality versus diversity? (iii) What social ramifications will repeated data loops have on information dissemination? 

To bridge this gap, in this research, we propose
\textbf{\underline{M}}odel 
\textsc{A}ut\textbf{\underline{O}}phagy 
\textsc{a}\textbf{\underline{NAL}}ysis (\OurMODEL{}) framework for explicating self-consumption within large models.
There exist several \textbf{motivations} underlying~\OurMODEL{}.
\textit{\underline{Firstly}}, large models are extensively being utilized across various domains~\citep{kaddour2023challenges}, and even crowd-sourced annotators heavily rely on generative AI for data curation and decision-making~\citep{veselovsky2023artificial}. 
\textit{\underline{Secondly}}, with internet being a direct source of 
    training data, the contemporary models are unwittingly being trained on AI-synthesized data~\citep{alemohammad2023selfconsuming, 
    Shumailov2023TheCO, veselovsky2023artificial}.
\textit{\underline{Thirdly}}, numerous studies opt to use the models as the generators and selectors of their training data, aiming to reduce overall training costs~\citep{li2023selfalignment, huang2023large}. 
This trend strictly implies that emerging large models are predominantly being 
trained on synthetic data, subsequently shaping subsequent human endeavors upon this synthetic foundation. We term this phenomenon as autophagy ("self-consumption").

\OurMODEL{} employs the concept of autophagous loops to analyze and comprehend the flow of information.
Specifically, it introduces two distinct variants of autophagous loops based on how humans and large models construct and utilize data or information from their surroundings, as illustrated in 
Figure~\ref{fig:frameworklms} and~\ref{fig:framework_human}, respectively. 
For evaluation,~\OurMODEL{} conducts comprehensive analytical and 
empirical analysis on a wide range of large models under both 
image and text data settings. 
These include (i) ``Cross-scoring experiments,'' designed to elucidate how humans and LLMs evaluate each other's responses; (ii) ``Exam scenario simulation,'' aimed at discerning the preferences of humans and LLMs in evaluating and filtering information;
(ii) ``AI-washing,'' demonstrating how generative models analyze, modify, and transmit information in a cyclical process. In experiments, we curated specialized datasets comprising text and images, ensuring a rigorous performance evaluation. 

\OurMODEL{}'s \textbf{findings} highlight:
\begin{enumerate}
    \item Large models tend to overrate their own answers while under-valuing 
    human responses, which clearly indicates that synthesized data is more 
    likely to prevail in information filtration processes.
    \item For each cycle of information exchange between humans and large models, 
    these models exhibit distinct preferences in amplifying or suppressing certain features. 
    This behavior not only hinders performance enhancements but also complicates 
    human intervention in the model's generative processes and information 
    transmission.
    \item It is worth noting that without ensuring a consistent presence of 
    real human-generated data, large models may increasingly rely on 
    self-generated datasets. This results in stagnating model performance. 
    We term this phenomenon as the large model converging to a 
    ``local optimum'', as elucidated in Section~\ref{sec:res_analyses}
\end{enumerate}

The rest of the paper is organized as follows. In Section~\ref{sec:related-work}, we provide related work. In Section~\ref{sec:background}, we introduce notations and offer a brief background. In Section~\ref{sec:proposed}, we introduce~\OurMODEL{}.
This is succeeded by comprehensive experimental evaluation and analyses in Section~\ref{sec:Experimentation}.
Finally, we conclude our findings in Section~\ref{sec:conclusion}.

\eat{To summarize, this paper we propose~\OurMODEL{} that uses autophagous 
loops to concentrate on the roles of large models as creators and distributors of information. 
\OurMODEL{} conducts a wide range of experiments to understand and 
evaluate the risks posed by the Human-AI interactions in terms of information quality and diversity, and how growing prevalence of
synthetic data may hinder performance improvements for 
next-generation/emergent LLMs.}

\eat{
This causes a growing prevalence of \warn{AI-generated} synthetic data within within human society as well as model training datasets.
The following summarizes the key contributions and findings of this paper:
To re-emphasize, Also
We explore how they handle data from various sources, each with unique characteristics, and examine how this data is either augmented or suppressed.}


\eat{,the widespread use of large models and}

\eat{Existing research has highlights key issues such as discrimination~\cite{Navigli3597307}, hallucinatory outputs~\cite{huang2023survey,tonmoy2024comprehensive}, and lack of interpretability~\cite{zhao2023explainability}.}

\eat{\textbf{\underline{M}}odel \textbf{\underline{A}}utophagy \textsc{a}\textbf{\underline{N}}alys\textbf{\underline{I}}s 
for self-\textbf{\underline{C}}onsumption 
\textsc{\textbf{Expla}}nat\textsc{\textbf{i}}o\textsc{\textbf{n}} (\OurMODEL{}).}

\eat{including but not limited to language 
and visual information, and multiple modes of interaction.}

\eat{whose properties are yet to be explored.
Autophagy is a technique widely used}

\section{Related Work}
\label{sec:related-work}
\noindent {\bf Generative AI for information production and dissemination.}
With widespread generative models, such as ChatGPT and DALL-E 3~\citep{betker2023dalle} etc., anyone can interact with AI
\eat{issue commands to AI} using natural language to express 
their feelings and/or posit different requirements. 
The AI uses multiple models to understand the content
followed by utilizing various resources to generate and 
curate information~\citep{kaddour2023challenges, yin2023survey}, 
which is later \eat{rapidly} disseminated through the internet. 
This break-through has significantly 
altered the role of AI in human society. 
Generative models are no longer just simple tools, rather 
they have become a crucial component in the production and 
dissemination of information~\citep{goldstein2023generative}. 

We emphasize that the risks associated with generative AI are 
not solely due to the biases and hallucination~\citep{huang2023survey, shen2023large}, \eat{which we have continually emphasized.}
they also stem from how humans interact with these systems, and 
the potential consequences, e.g., the creation of "information cocoons"~\citep{piao2023humanai}.

\noindent {\bf Self-Training and Self-Consuming \eat{Large }Models.} 
Recently, there has been a surge in use of automated routines/models 
for model alignment, data filtering, and data enhancements
~\citep{gulcehre2023reinforced, li2023selfalignment}, helpful to
avoid the significant costs associated with creating 
humanly annotated data sets.
A large number of LLMs-generated datasets are being 
used to fine-tune pre-trained foundation models~\citep{alpaca, instructionwild}.
Simultaneously, the most powerful models currently available 
are often used as judges in numerous competitions~\citep{vicuna2023}. 
The risk involved in these approaches is significant, as models 
have already been preliminarily proven to be biased rather than 
\eat{objective and} impartial~\citep{wu2023style,liu2023llms}. 
~\citep{alemohammad2023selfconsuming} proposed an autophagous 
loop for the computer vision models. Their work, characterized 
by the models trained using data generated by the models 
themselves, led to a decline in model performance and data 
diversity~\citep{Shumailov2023TheCO}. 
Subsequent studies have demonstrated similar traits in  
language models~\citep{briesch2023large}.

\eat{Simultaneously, a large number of datasets~\citep{alpaca, instructionwild} generated by LLMs are also used to fine-tune pre-training foundation models, and the most powerful models currently available are often used as judges in "model competitions"~\citep{vicuna2023}. The risk involved in these approaches is significant, as models have already been preliminarily 
proven to be less than objective and impartial~\citep{wu2023style,liu2023llms}.}

\section{Background/Preliminaries}
\label{sec:background}
\paragraph{Notations:}
In this paper, we use
$d$ to represent the domain of QA-pair, helpful for providing 
question-specific context;
$Q$ to represent the question;
$D$ to represent doc-level information helpful for answering $Q$;
$A$ to represent the human response/answer;
$A_{m_i}$ to represent the response generated by model $m_{i}$, 
with $i \in \{1,6\}$ represent six different LLMs;
$A_{\text{m}_i\text{s}_5}$ to represent the highest quality 
response by model $m_i$;
and $A_{\text{m}_i\text{s}_1}$ to represent the lowest quality 
response by model $m_i$.

\eat{
In the context of our dataset, the following notation is used to represent elements within a tuple and the answers generated by various language models:
\begin{itemize}
    \item $ d $: The domain of the question-answer pair, providing context for the question classification.
    \item $ Q $: The question posed by a user, serving as direct input for model-generated answers.
    \item $ D $: Document-related information necessary for answering $ Q $ with background knowledge.
    \item $ A $: The human answer that received the most endorsements for question $ Q $, used as a benchmark for answer quality.
\end{itemize}    
For each language model \( \text{model}_i \), where \( i \) ranges from 0 to 5, representing one of six different large language models:
\warn{
\begin{itemize}
    \item \( A_{\text{m}_i} \): The initial answer generated by model \( i \).
    \item \( A_{\text{m}_i\text{s}_5} \): The highest quality answer generated by model \( i \), according to prompts.
    \item \( A_{\text{m}_i\text{s}_1} \): The lowest quality answer generated by model \( i \), according to prompts.
\end{itemize}
}}

\subsection{Autophagous Loops} 
\label{sec:autop-def}
\begin{figure*}[t]
    \centering
    \begin{minipage}{0.45\linewidth}
        \centering
        \includegraphics[width=\textwidth]{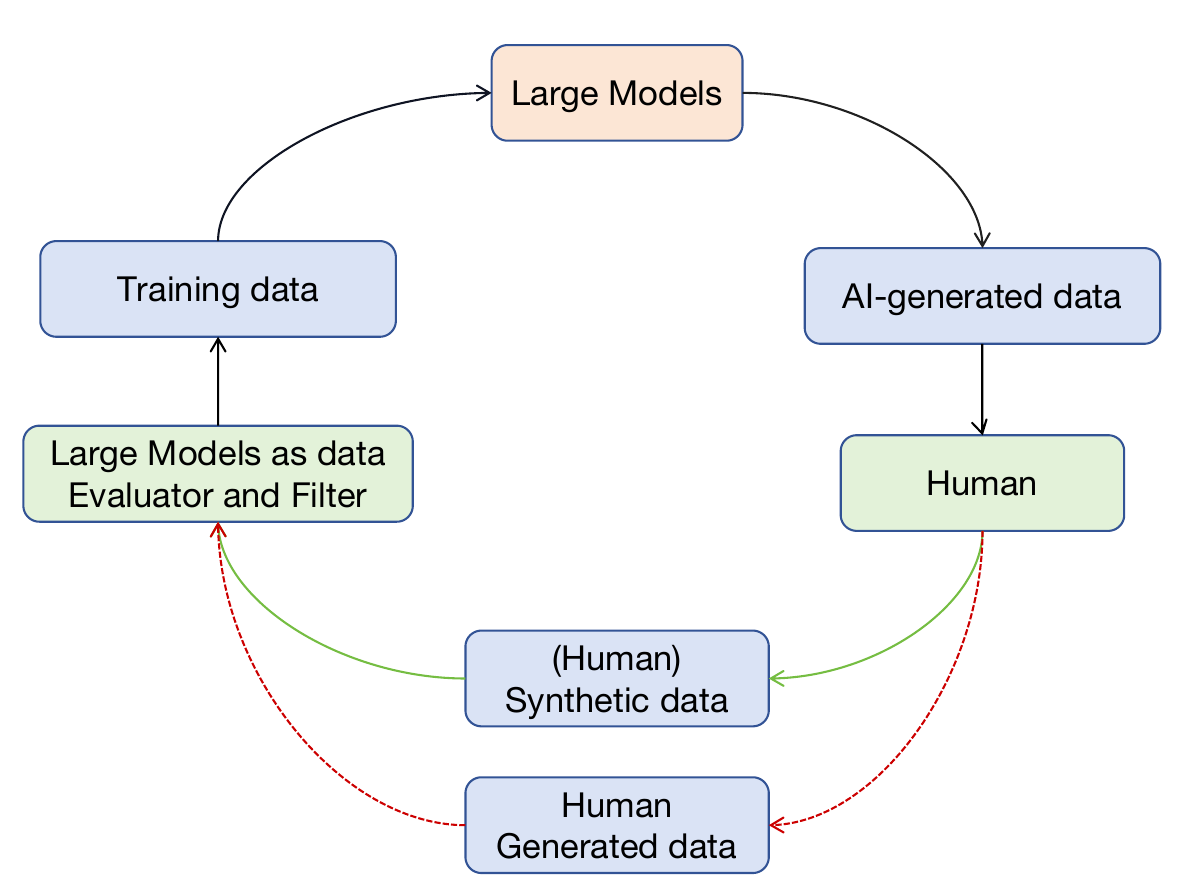}
        \caption{Self-consumption loop of large models. This figure is based on recent workflows for automated data generation and filtering \citep{wang2023selfinstruct,li2023selfalignment}. We emphasize the preferential nature of large models as generators and filters of synthetic data.}
        \label{fig:frameworklms}
    \end{minipage}
    \hfill
    \begin{minipage}{0.45\linewidth}
        \centering
        \includegraphics[width=\textwidth]{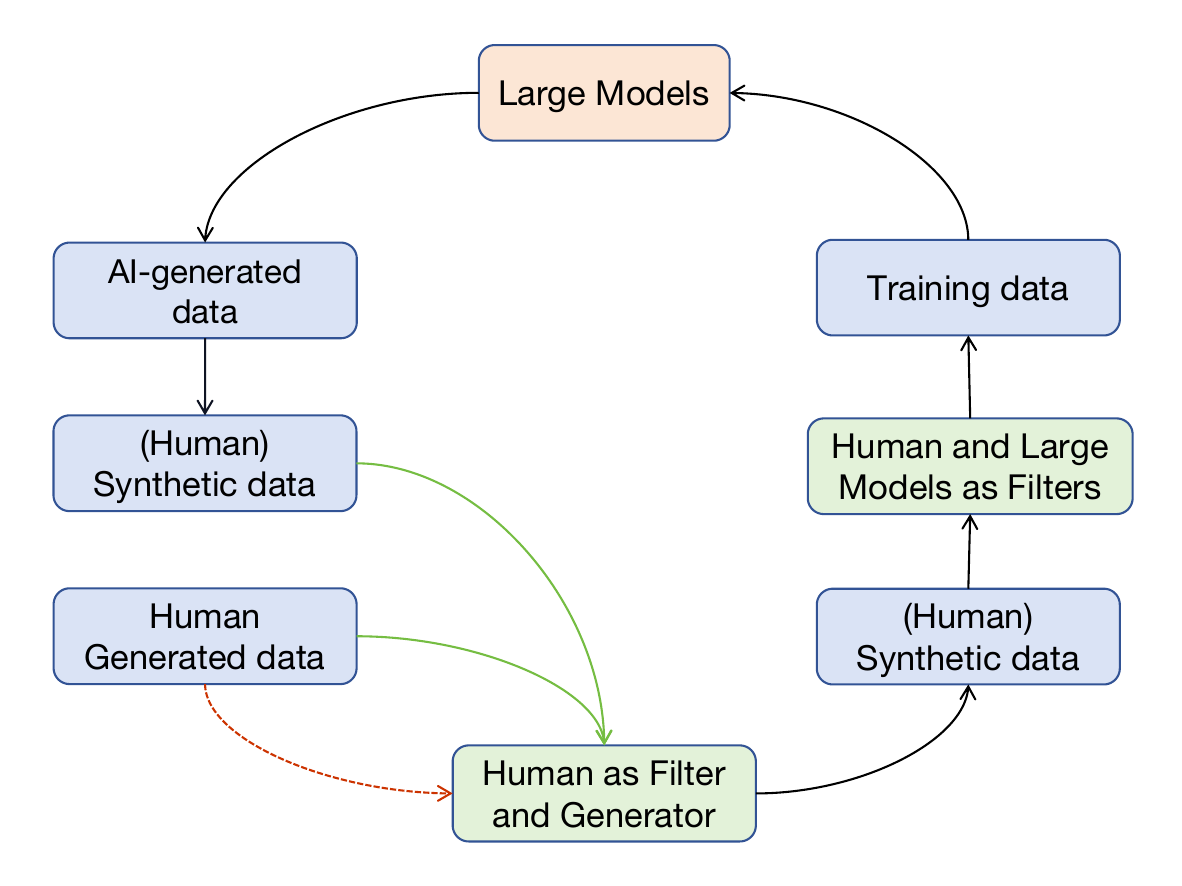}
        \caption{Self-consumption loop emphasizes the role of humans as filters and transmitters of information~\citep{veselovsky2023artificial} while interacting with large models. Such a role primarily exists during the process of information dissemination in human society. }
        \label{fig:framework_human}
    \end{minipage}
\vspace{-2.7ex}
\end{figure*}

We redefine the relationship
between large models and human societal information dissemination
by drawing inspiration from the classic communication theory
of the Ritual view by~\citet{Carey2008Communication} (see Appendix \ref{sec:Ritual_view} for details).
\eat{Drawing from the classic communication theory of the Ritual view as 
proposed in \citep{Carey2008Communication}, we redefine the relationship
between large models and human societal information dissemination (see Appendix \ref{sec:Ritual_view} for details).}
This is illustrated in \eat{As shown in }Figures \ref{fig:frameworklms} 
and~\ref{fig:framework_human}, where we show both large models and humans can act as generators and filters of information in the Human-AI communication system.
However, this system is prompting machine learning algorithms
to encode all the stereotypes, inequalities, and power
asymmetries that exist in human society~\citep{unseeblackface}.
For instance, women with darker skin are more likely to be 
misclassified in gender classification compared to men with 
lighter skin, which is due to the majority of samples in the 
training datasets being subjects with lighter skin tones
\citep{buolamwini2018gender}. 
The biased information generation and transmission process of 
large models and humans will further exacerbate these phenomena.

\section{\OurMODEL{}}
\label{sec:proposed}
\vspace{-1.7ex}

In this work, We propose~\textbf{\underline{M}}odel 
\textsc{A}ut\textbf{\underline{O}}phagy \textsc{a}\textbf{\underline{NAL}}ysis (\OurMODEL{}) 
for large models' self-consumption explanation.
The core contributions of~\OurMODEL{} include:
(1) designing realistic models of autophagy by drawing
conclusions from human behaviors in utilizing large models;
(2) curation of novel datasets (both text and image) followed 
by rigorous experimental studies to demonstrate/showcase how real-world 
data distribution is influenced by large models.

\eat{\subsection{\warn{Realistic Models} for Autophagous ("self-consuming") Loops}}

\subsection{Self-consumption Loops}
\OurMODEL{} uses two different models for ``self-consumption" loops 
(relying on autophagous loops) to simulate the
interaction between humans and large models.
We argue unlike previous work~\citep{alemohammad2023selfconsuming},
\OurMODEL{} offers a more realistic and natural setting.

The end goal of~\OurMODEL{} is to understand different biases incurred 
by using humans and/or LLMs as generators and {transmitters} of 
information. These biases could help us understand the loss in 
data quality and/or diversity for LLM-generated and human societal 
datasets. It also explicates the role these models play in the 
exchange of information within human society from a broader perspective.
The self-consumption loops of large models and humans employed 
by~\OurMODEL{} are explained as follows.

\noindent \textbf{Large Models.}
Figure~\ref{fig:frameworklms} outlines the cyclical influence 
of large models in the data processing life-cycle.
As demonstrated in the figure, the training data undergoes a 
transformation through either algorithmic refinement and/or 
human curation, resulting in what we term ``synthetic data", 
see the ``AI-generated data - Human - Synthetic data pathway"
(shown as input to large models via solid green color).
This contrasts with the ``human-generated data", which primarily 
originates directly from humans and is typically less 
structured (shown as input to large models via a dotted red line). 

We claim that large models that have access to both types of 
datasets are usually biased to filter preferentially, 
and/or prefer synthetic data over human-generated data for 
future learning cycles.
This bias may either be inherently {preferential} or is 
incurred by the model training objective. To empirically validate 
this claim, we report multiple experiments in 
Section~\ref{sec:Experimentation}.

Furthermore, Figure~\ref{fig:frameworklms} also explicates 
that the human's role in this cycle is not entirely passive. 
Humans, influenced by the outputs of large models, may
unknowingly prioritize synthetic data due to its 
processed nature, which seems more immediately usable or 
relevant~\citep{veselovsky2023artificial}. 
This preferential feedback loop can inadvertently lead to 
the diminishing the raw and/or human-generated data in the 
pool of data resources for being perceived as less refined.

\noindent\textbf{Humans.} 
Figure~\ref{fig:framework_human} showcases the specific 
behaviors of humans when interacting 
with large models. It presents a more detailed view for 
human-agent interactions, highlighting the fact\eat{empirical finding} 
that humans tend to favor the data generated by large models.
This is illustrated in the Figure by using solid green lines 
showing high preference compared to the red dotted line, indicating 
inhibition.
This fact/claim is also validated by our later rigorous experimentation,
which emphasizes that without transparent data provenance, humans 
may prefer outputs of large models, thus contributing to the 
cyclical bias toward synthetic data. 

Correlating these two figures, we infer that the 
relationship between these two loops is symbiotic, offering a 
microcosm of humans' preferences in the Human-AI communication.

\eat{Correlating two Figures, we infer that the 
relationship between these two loops is symbiotic; while 
Figure~\ref{fig:frameworklms} provides an overview of the data 
cycle in a large model training loop, Figure~\ref{fig:framework_human} 
zooms in on the human aspect, offering a microcosm of human 
preference in the Human-AI communication.}

\subsection{Rationality and Risks \eat{of Autophagous  Loops}}
\label{sec:rat_risk}

In order to understand the rationale of the self-consumption loops on the data quality vs. diversity and examine the risks imposed,~\OurMODEL{} performs comprehensive experimentation.
Note that a core proposition of our proposed models is that the 
large models and humans cannot maintain objectivity and impartiality as part of the information dissemination loop, which could 
also be indicated by the colored line segments in 
Figure~\ref{fig:frameworklms} and~\ref{fig:framework_human}. 
In order to validate this proposition,~\OurMODEL{} performs the following 
different analyses:

\eat{, with dotted-red representing inhibition 
and solid green representing facilitation.
The details of the analysis are as follows:
}

\noindent \textbf{Cross-Scoring Experiment.}
Cross-scoring experiments aim to demonstrate the inhibitory and promotive phenomenons of information transmission within autophagous loops.
We focus on whether LLMs and humans remain impartial while filtering and transmitting information and, if not, what kind of bias they induce.
For this, we employ mainstream LLMs to generate question-answer pairs 
based on prompts 
(in Appendix~\ref{appendix:Generation_Prompt_Template}), and 
instructed them to perform cross-scoring, i.e., using one model 
to evaluate and assign scores to the response generated by 
other models, and so on.

In order to mitigate the impact of specified scoring 
standards, i.e., tenths, percentages, and specific rules explained 
in Appendix~\ref{appendix:Cross_scoring_Prompt}, 
we designed a simulated testing scenario to analyze which 
is more likely to prevail in the cycle of information 
dissemination in real-world scenarios: humanly-generated 
or AI-generated answers. This analysis aims to examine the consistency and 
the bias of humans and language models in adhering to these scoring 
standards.

\noindent {\bf Exam Scenario Simulation.}
Exam scenario simulation aims to 
answer the following question: Human-generated or AI-generated 
answers, which one wins in information screening and filtering?
For this, we use LLMs and humans to simultaneously act as the generators 
and evaluators, see Section~\ref{fig:Exam_Scenario_Simulation} for process-flow.

In this simulation scenario, the answers 
generated by LLMs alongside those produced by humans, are 
anonymized to mitigate any bias.
To further eliminate the potential influence of the sequence 
in which the answers are presented, we also randomized their 
order. Both language models and human participants, assuming the role 
of experts were then asked to assign scores to these 
answers on a percentile scale and chose the best answer.

\noindent \textbf{AI-washing.} Finally, we conducted 
an ``AI-washing" experiment in order to explore the risks 
posed by large models and humans as information 
generators, and to observe the changes in real 
data after multiple rounds of AI refinements. 
For these experiments, our goal is to analyze the trade-off between 
the information quality vs. diversity and comprehend 
large models' ability to enhance and weaken different 
information contents, i.e., whether large models 
are faithful messengers of information?

Basically, we aim to answer the following questions:
(i) Do these models capture the main points in the information that 
needs to be conveyed? 
(ii) Whether large models play the role of a link in the information 
transmission that may also lead to losses?


\vspace{-0.1in}
\section{Experimentation}
\label{sec:Experimentation}
\vspace{-0.1in} 
In this section, we perform comprehensive experimentation
of~\OurMODEL{} in order to evaluate the preferences of humans and language 
models in information selection. Experimental details are as follows.
\eat{This helps us analyze how real human-generated data 
is suppressed in the human vs language model 
information dissemination loop.
The core evaluation objective is to:
(i) examine the preferences of both humans and LLMs in 
evaluation and filtration of information as part of the 
dissemination process;
(ii) investigated the potential drawbacks associated 
with using generative models for enhancing and transferring 
information.}

\vspace{-0.1in}
\subsection{Experimental Settings}
\label{sec:exp-settings}
\vspace{-0.1in} 
\noindent {\bf Datasets.}
For experimental evaluation, we curated three different 
datasets, namely:
(i) QA-pairs,
(ii) Book3,
(iii) Image-ax.
QA-pairs is a structured text dataset used for cross-scoring and 
exam scenario simulation.
Book3 and Image-ax are unstructured datasets used for AI-washing 
experiments for text and images, respectively.
Detailed description and statistics of the dataset is provided in
Appendix~\ref{appendix:Dataset}.

\noindent {\bf Large Models.}
We employed six different LLMs with varying architectures.
These include 
ChatGPT~\citep{li-etal-2022-gpt}, 
GPT-4~\citep{openai2023gpt4}, Claud2\footnote{\url{https://www.anthropic.com/index/claude-2}}, 
Llama-2-70b-chat~\citep{touvron2023llama}, 
PaLM2-chat-bison\footnote{\url{https://blog.google/technology/ai/google-palm-2-ai-large-language-model/}}, and 
Solar-0-70b-16bit\footnote{\url{https://huggingface.co/upstage/SOLAR-0-70b-16bit}}. 
We opted for relatively large-scale models owing to their 
superior capabilities in instruction adherence, which we found 
lacking in small-scale models.
\eat{The focus on larger models in our experiments is due to their 
superior capability in instruction adherence and context length 
handling, which we found lacking in smaller-scale models.\fixme{Move it.}}
In assessing textual diversity, we used the models 
bge-large-zh-v1.5 and bge-large-en-v1.5
by~\citet{bge_embedding} as the embedding models.
For computer vision tasks, we utilized the open-source model 
StableDiffusionXL (SDXL) by~\citet{podell2023sdxl}.

\noindent {\bf Evaluation Metrics.}
For cross-scoring and exam simulation experiments, we 
alternatively use humans and LLMs as response generators and/or evaluators. 
The corresponding template for using LLMs as evaluators is given in 
Appendix~\ref{appendix:Cross_scoring_Prompt}.
This scoring criteria aims to verify the objective nature of LLMs 
as evaluators, i.e., whether a model assigns a higher 
scores to their response or response generated by other 
models.
For AI-washing experiments, we measure the diversity in information
after multiple rounds of iterations.
Specifically, for the text data, we use the cosine 
similarity (shown in Appendix~\ref{Appendix:datavec-sim}) 
to measure the diversity in original text and text 
reproduced by LLMs.
For the image dataset, we visually analyze which features 
are preserved, omitted, and transformed by the LLMs.

\noindent {\bf Experimental Setup.}
For AI-washing experiments, the number of iteration round $N$ 
for text and images is 20. 
However, we find that for text, the large model will no 
longer make significant changes to the text when $N > 4$. 
For images, however, the large model performs significantly 
differently across samples, as we discuss in detail later.

\vspace{-0.1in}
\subsection{Experimental Results and Analysis}
\label{sec:res_analyses}
\vspace{-0.1in} 

\subsubsection{Cross-scoring Experiment.}
\label{sec:cross-scoring}
The results of~\OurMODEL{} for the cross-scoring experiment for QA-pairs are
shown in Table~\ref{table:exp1}. For these results, we report 
the scores assigned by LLMs and human annotators under 
the cross-scoring setting.
\begin{table*}[!t]
    \centering
    \small
      \resizebox{\textwidth}{!}{%
   \begin{tabular}{lllllllll}
    \toprule 
        \textbf{Scorer / Generator} & \textbf{ChatGPT} &  \textbf{GPT4} & \textbf{Claud2} &  \textbf{Llama-2-70b-chat} &  \textbf{PaLM2-chat-bison} &  \textbf{Solar-0-70b-16bit} & \textbf{Human}   & \textbf{Average}\\
    \hline
    \multicolumn{9}{c}{\cellcolor{gray!25}\scshape Originally Generated Answer} \\
    \hline
        ChatGPT & \textbf{4.33}& \textbf{4.29}& 3.88& 4.25& 3.92& 4.17& \textbf{2.48}
 &3.90\\ 
        GPT4 & \textbf{4.63}& \textbf{4.56}& 4.04& 4.41& 3.95& 4.60& \textbf{2.77}
 &4.14\\ 
        Claud2 & 3.92& 3.97& 4.00& 4.00& 3.95& 3.97
& \textbf{3.36}
 &3.88\\ 
        Llama-2-70b-chat & 3.91& 3.99& 3.82& 4.00& 3.61& 3.90& \textbf{3.23}
 &3.78\\ 
        PaLM2-chat-bison & 3.99& 4.05& 3.72& 4.22& 3.60& 3.77& \textbf{3.57}
 &3.85\\ 
        Solar-0-70b-16bit & 4.10& 4.35& 4.05& 4.16& 4.01& 4.12
& \textbf{2.59}
 &3.91\\
 Human& 4.75& 4.79& 4.50& 4.18& 4.28& 4.17&\textbf{3.58} & \textbf{4.32}\\ 
    \hline
        \multicolumn{9}{c}{\cellcolor{gray!25}\scshape Best Quality Answer} \\ 
        \hline
        ChatGPT & 4.24& 4.28& 4.41& 3.80& 4.21& 4.20
& - &4.19\\ 
        GPT4 & \textbf{4.52}& \textbf{4.75}& 4.20& 4.11& 4.00& 4.36& -
 &4.32\\ 
        Claud2 & 3.92& 3.98& 4.21& 4.20& 4.01& 3.97
& -
 &4.04\\ 
        Llama-2-70b-chat & 3.91& 4.03& 4.26& 4.07& 4.30& 3.95
& -
 &4.09\\ 
        PaLM2-chat-bison & 3.98& 4.23& 4.42& 3.84& 4.26& 3.98
& -
 &4.12\\ 
        Solar-0-70b-16bit & 4.34& 4.43& 4.42& 4.33& 4.28& 4.11
& -
 &4.32\\ 
        Human & 4.23& \textbf{4.92}& 4.30& 4.20& 4.07& 4.26& -
 &\textbf{4.33}\\
         \hline
        \multicolumn{9}{c}{\cellcolor{gray!25}\scshape Worst Quality Answer} \\ 
         \hline
        ChatGPT & \textbf{3.13}& 1.33& 1.27& 1.27& 2.83& 2.21
& - &2.01\\ 
        GPT4 & \textbf{3.19}& 1.40& 1.29& 1.33& 2.98& 1.70& - &1.98\\ 
        Claud2 & \textbf{4.08}& \textbf{3.23}& \textbf{3.71}& 1.76& \textbf{3.85}& \textbf{3.77}
& - &3.40\\ 
        Llama-2-70b-chat & 2.69& 1.06& 2.17& 1.78& 2.27& 2.11
& - &2.01\\ 
        PaLM2-chat-bison & 2.65& 1.23& 1.28& 1.69& 2.73& 2.31
& - &1.98\\ 
        Solar-0-70b-16bit & \textbf{3.28}& 1.26& 1.89& 2.40& 2.37& 2.40
& - &2.27\\ 
 Human& 1.76& 2.31& 1.24& 1.33& 2.00& 1.82&- &\textbf{2.09}\\
  \bottomrule
    \end{tabular}%
}
\caption{Cross-scoring analysis of language models and humans. 
The results in this table are average scores out of a five-point scale, assigned by both models and human evaluators, to the generated answers. 
These scores are calculated based on the criteria outlined in 
 Appendix~\ref{appendix:Cross_scoring_Prompt}, and
Appendix~\ref{appendix:Scoring Criteria for Human Evaluation}. 
The table is organized into three sections: with 
(i) ``Originally Generated Answer'', representing the scores for 
original response;
(ii) ``Best Quality Answer", representing scores $(A_{\text{m}_i\text{s}_5})$; 
and (iii) ``Worst Quality Answer", representing scores 
$(A_{\text{m}_i\text{s}_{1}})$. We bold-face the best scores in each section.}
\label{table:exp1}
\vspace{-12pt}
\end{table*}
For scoring via LLMs, we prompt each model to assess not only 
the answers generated by other models but also those produced by humans. 
The scoring range is between 1 to 5 (see 
Appendix~\ref{table:The prompt template for evaluating answers}),
with 1 indicating low quality and 5 for best quality answer. 
For human evaluation, we employed fifty crowd-sourced annotators 
to rate all question-answer pairs based on the 
scoring criteria in 
Appendix~\ref{table:Scoring criteria for crowdsourced annotations}. 
{Note, we recorded the average scores for all samples, 
excluding instances where the LLMs and/or human evaluators 
refused to respond.}  These results indicate:

{\bf Biasedness in Information Selection.}
These results help us understand that
LLMs do have an inherent capability to comprehend 
specific criteria of the prompt in Appendix~\ref{appendix:Cross_scoring_Prompt}
and can adjust the quality of their generated answers based 
on relevant instructions.

However, we observe that each model exhibits certain preferences.
Specifically, models tend to assign higher scores to the high-quality 
answers generated by themselves, particularly for ChatGPT and GPT-4, 
which both demonstrate high confidence in their own outputs. 
Also, we find that 
ChatGPT and GPT-4 exhibit similar characteristics in scoring; 
they are inclined to extreme scores, i.e., high or low scores 
(assigning scores of 1 or 5 with very high probability), 
at the same time. They both tend to 
assign lower scores to Claude2 and PaLM2-chat-bison.
These results are aligned with earlier research by~\citet{liu2023llms} 
that emphasized that current top-performing LLMs 
(both black-box and white-box) are narcissistic evaluators.

While evaluating the abilities of LLMs as answer/response generators, 
ChatGPT’s worst-quality answers, which are easy to figure out for humans 
via crowd-sourcing,  can still deceptively obtain 
higher scores from other models. Furthermore, we observed that Claud2 tends to favor neutral 
and less-controversial ratings, often assigning average 
scores, i.e., 3 or 4, even for the worst quality answers. 
This is also reflected by a relatively higher score assigned 
by Claud2 to low-quality answers compared to other models. 
\eat{is much higher than that of other models,} 
Also, the difference between its score for initial answers and best-quality answers is not significant.
Llama-2-70b-chat and Solar-0-70B-16bit exhibit similar scoring 
and generation behaviors, which may be attributable to the fact
that Solar-0-70B-16bit is fine-tuned form LLaMA-2-70b 
with only Orca-style~\citep{mukherjee2023orca,OpenOrca} 
and Alpaca-style dataset~\citep{alpaca}, indicating that the 
pre-trained model has a significant influence on the model's 
preferences.

We also observed that prompting the model to generate better answers does  
not always lead to higher scores in our experiments. Only Claud2 
showed significant improvement in the answers compared to the 
original responses. Conversely, when we prompt model to generate 
poorer answers, the quality of answers generated by almost all 
LLMs in Table~\ref{table:exp1} decreased significantly.
However, ChatGPT and Palm2-chat-bision still achieved relatively 
high scores, a possible reason could be attributed to the models 
being highly aligned to avoid producing harmful 
outputs~\citep{lambert2022illustrating}, so irrespective of input 
prompts, they always try to positively answer the question. 
We leave further investigation as a future research direction.

Furthermore, the scoring behavior of crowd-sourced annotators toward the answers generated by the large language model was largely consistent. This may be due to the highly structured and standardized nature of answers produced by AI systems aligned with human feedback. In contrast, human-generated answers received lower scores from the human annotators, 
with significant variations among different evaluators. This may be attributed to the higher degree of alignment between the large model and the human collective compared to alignment among individual humans.



\begin{table}[b]
\vspace{-2.1ex}
    \centering
    \small
    \resizebox{0.45\linewidth}{!}{
    \begin{tabular}{@{}lccc@{}}
    \toprule
    & \multicolumn{3}{c}{Evaluator} \\ 
    \cmidrule{2-4} 
    Generator & ChatGPT & Claud2 & Human \\
    \hline
    \multicolumn{4}{c}{\cellcolor{gray!25}\scshape Average Score} \\
    \hline
    ChatGPT & \textbf{95} & \textbf{90} & \textbf{91.7} \\
    Claud2  & 92 & 88 & 90 \\
    Human   & 90 & 75 & 80 \\
    \hline
    \multicolumn{4}{c}{\cellcolor{gray!25}\scshape Selected as Best Answer} \\
    \hline
    ChatGPT & 41 & \textbf{58} & \textbf{66 }\\
    Claud2  & \textbf{55} & 42 & 25 \\
    Human   & 4  & 0  & 9 \\
    \bottomrule
    \end{tabular}}
     \caption{Result for exam scenario simulation. We boldface the best scores.}
     \label{table:Exam_Scenario_Simulation}
\vspace{-12pt}
\end{table} 
\subsubsection{Exam Scenario Simulation.}
\label{sec:exam_scenario}
The exam simulation scenario aims to mitigate the impact of scoring 
criteria on the evaluative capabilities of large models and 
human evaluators in the previous experiments. It helps to understand the preferences 
of LLMs and humans in evaluating and filtering information 
(see Appendix~\ref{fig:Exam_Scenario_Simulation} for process-flow).
For this experiment, we used LLMs (ChatGPT, Claud2) and humans to 
simultaneously act as generators and evaluators.
The results of~\OurMODEL{} for exam scenario simulation are shown in 
Table~\ref{table:Exam_Scenario_Simulation}.
We report the average scores assigned by evaluators for the cases 
when their responses were selected as the best answers. 
These results show:

{\bf LLMs win in information screening.}
Responses from LLMs consistently garnered acceptance from 
evaluators and were frequently chosen as the best answers. 
In contrast, human-generated responses were rarely chosen 
as the best response, indicating a challenge in integrating 
authentic human-generated responses into models' training 
data and the real-world human feedback loop.

This finding substantiates the potential risks also highlighted 
in Section~\ref{sec:rat_risk} that emphasized that human-generated 
responses typically receive comparatively lower consideration 
in the self-consumption loops.

 \begin{figure*}[t]
    \centering
    \begin{minipage}{0.53\linewidth}
        \centering
        \includegraphics[width=\textwidth]{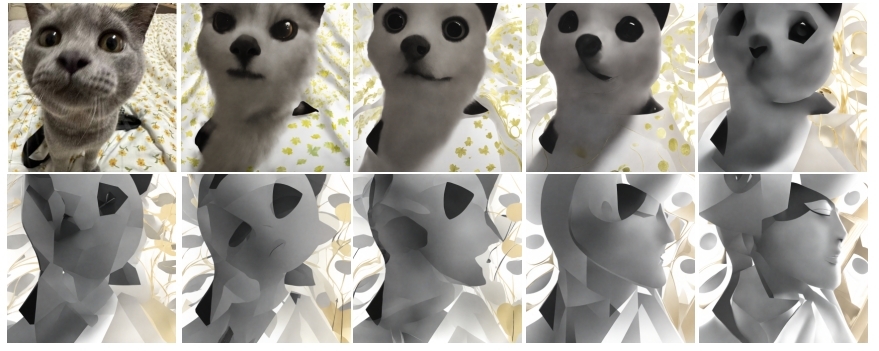} 
        \caption{An example illustration of AI-washing on images that shows that repeatedly processing images $N$-times ($N$=1:5) using SDXL model~\citep{podell2023sdxl} may lead to serious biases. }
        \label{fig:SDXL_0.png} 
    \end{minipage}
    \hfill
    \begin{minipage}{0.45\linewidth}
        \centering
        \includegraphics[width=\textwidth]{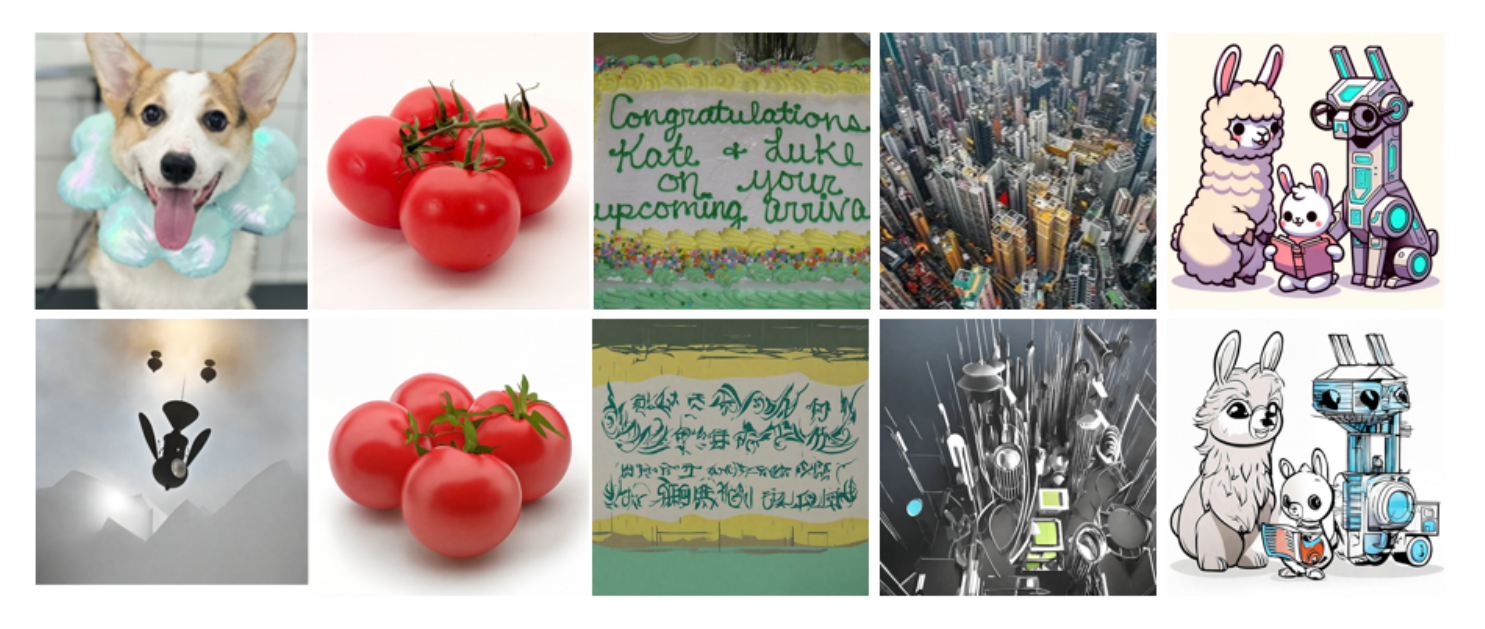}
        \caption{After 20 rounds of AI-washing experiments with the SDXL model~\citep{podell2023sdxl}, it becomes evident that different images retain and discard details in varying manners. }
        \label{fig:aiwashimag}
    \end{minipage}
\vspace{-2.7ex}
\end{figure*}
\subsubsection{\bf AI-washing.} 
\label{sec:AI_Washing}

Example demonstration of AI-washing experiments for 
text data set processed via ChatGPT are shown in 
Appendix~\ref{Appendix:ai-washtext} (Table~\ref{table:aiwashtext}).
The prompts used for these experiments are 
provided in Appendix~\ref{appendix:C}. 
Likewise, some examples of AI-washing experiments for images are 
shown in Figure~\ref{fig:SDXL_0.png} and~\ref{fig:aiwashimag}.
For this, we instruct SDXL~\citep{podell2023sdxl} to process 
these samples $N$ times. These results show:

{\bf LLMs are Biased Information Transmitters.}
Large models are inherently biased regarding the manner and content of 
conveyed information.
For instance, textual example (in Table~\ref{table:aiwashtext}) 
processed $N$ times ($N$=1:5) by ChatGPT shows subtle shifts in the 
sample's language style and narrative technique.
Similarly, for images in Figure~\ref{fig:SDXL_0.png}, 
the model preserved the color distribution of the original
image but altered the main subject from a cat to a human portrait. 
Conversely, for the images in Figure~\ref{fig:aiwashimag}, 
we observed a dominant alteration in image style rather than a change in the primary content for the first, fourth, and fifth images.
Whereas, the images at the second and third positions underwent 
several iterations with minimal transformation, with
the third image lost readability for the textual content.

This inconsistency may be attributed to the fact that while SDXL 
is renowned for generating high-quality images, the definition 
of quality in the context of generative models is subjective 
and heavily influenced by the annotations of the training dataset, also explained in detail by~\citet{podell2023sdxl}. 
These observations indicate a bias in the model's processing, 
where certain features are selectively preserved or altered based 
on the model's training objective and inherent design. 
This results in a higher probability of generating images 
containing specific features and/or information, e.g.,
loop, such as hand-drawn styles, portraits, close-ups of objects 
with clear backgrounds, etc.

To summarize, these findings demonstrate that large models are 
inherently biased regarding the manner and content of conveyed 
information. And repeated processing of images with 
generative models is akin to information and feature 
filtration, where generative models tend to emphasize 
or de-emphasize certain features.

\begin{figure*}[t]
    \centering
    \begin{minipage}{0.45\textwidth}
        \centering
        \includegraphics[width=\linewidth]{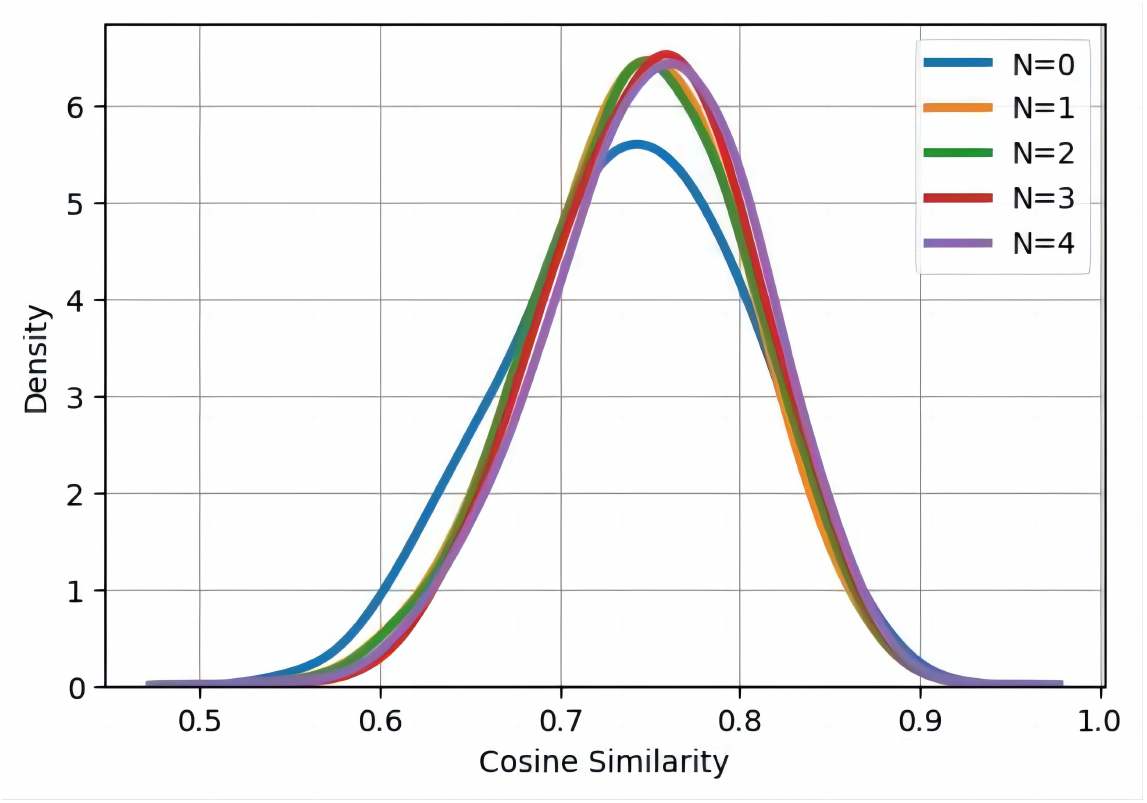}
        \caption{Density distributions of cosine similarity scores for text 
        samples from Book3 processed $N$ times by ChatGPT.}
        \label{fig:average_simchange}
    \end{minipage}
    \hfill
    \begin{minipage}{0.50\textwidth}
        \centering
        \includegraphics[width=\linewidth]{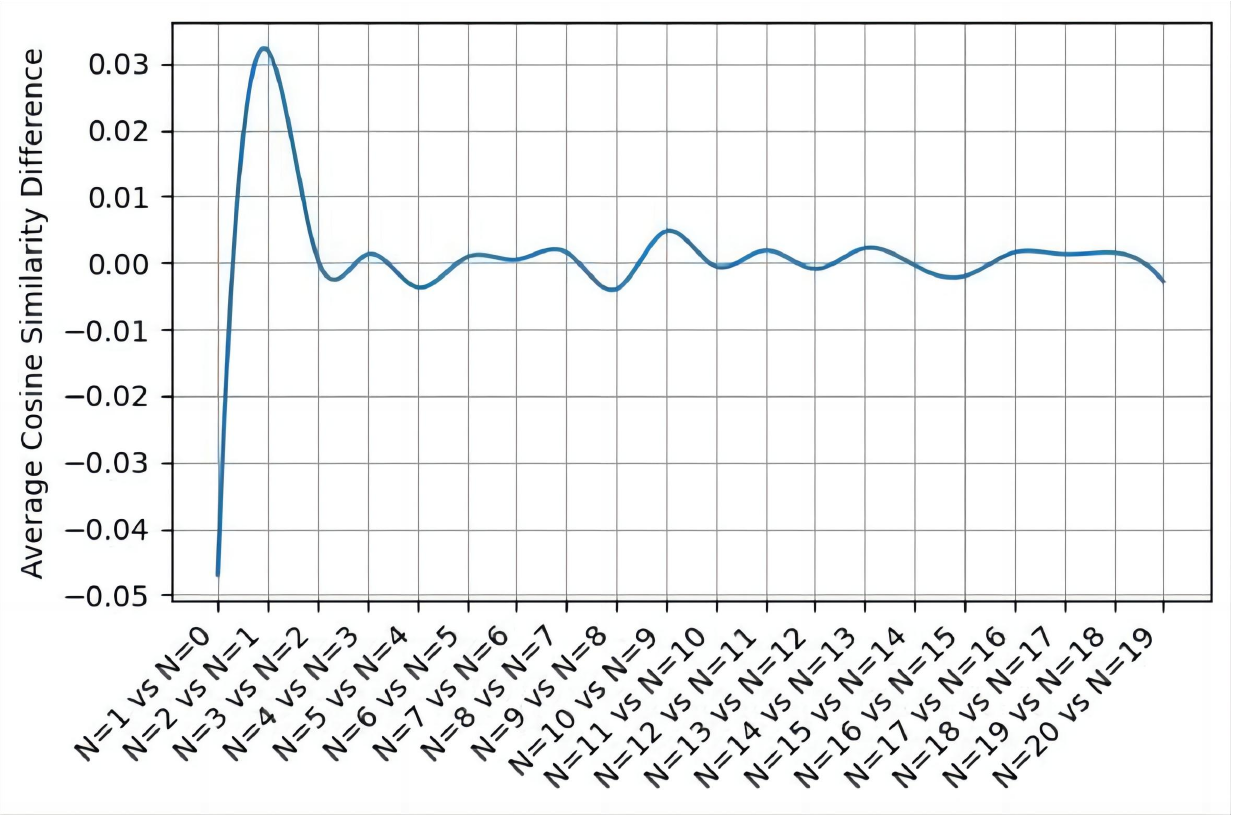}
        \caption{We report difference in cosine similarity for $N$ successive iterations 
        ($N\leq$20). For this graph, we use samples from Book3 dataset and Chat-GPT as LLM.  }
        \label{fig:average_distributionchange}
    \end{minipage}
\end{figure*}

{\bf Information Diversity.}
In order to further understand the impact of model training on data 
quality and diversity, we analyze the training behavior of large 
models. Specifically, we analyze the cosine similarity 
scores for multiple iterations of humanly-generated and LLM-generated text.
Corresponding results in Figure~\ref{fig:average_simchange} shows that
for successive rounds of text processed using LLMs, the cosine similarity 
between texts increases significantly, with the lower tail (i.e., 
line $(N = 0)$ with cosine similarity b/w 0.5 and 0.7) being washed out.
This shows that while acting as information disseminators, the 
large models exhibit some unique characteristics, i.e., they tend  to optimize more for real-world data while being more lenient 
towards self-generated samples.

We also analyzed the difference in cosine similarity across successive runs.
The line graph in Figure~\ref{fig:average_distributionchange} shows that
after approximately three iterations, the difference in the average 
cosine similarity across multiple rounds tends 
to stabilize. 
With the growing utility of large models, especially as a mechanism for knowledge/information comprehension, this 
phenomenon poses significant risks to the fairness and 
diversity of information dissemination.
\eat{We leave the specific details of this process to future research.}

{\bf Local Optimum.}
To re-emphasize the results in Figure~\ref{fig:average_distributionchange},
we observe that starting from the third iteration, the samples appear 
to have reached a ``local optimum", requiring fewer updates in model parameters.  
This highlights that: (i) By the third iteration, the large model has significantly 
adapted to the data details and/or style; 
(ii) While the LLM effectively extends its knowledge about the original content, it somehow falls short of achieving self-styling changes and updates. 
This self-bias suggests that when processing information, the large models 
learn to rewrite the text of different styles into a uniform style rather 
than enhancing the quality and diversity 
of the text also mentioned by~\citet{xu2024perils}.
This tendency could lead the model to become entrenched 
in its own style, potentially limiting its adaptability and creativity.



\eat{For instance, in the \warn{"five-score answers" segment, 
(Table~\ref{table:The prompt template for evaluating answers}}), 
an answer generated by ChatGPT would be evaluated by other LLMs like 
GPT4 and Claud2, assigning a score within the 1-5 range based on 
its quality.}

\eat{(Figure~\ref{fig:average_distributionchange}).
\warn{This indicates an 'alignment' process of the large 
models when processing information.}}
\vspace{-7pt}

\section{Conclusion}
\label{sec:conclusion}
In this work, we propose~\OurMODEL{} that uses two different self-consumption loops 
to examine large models as generators and disseminators of information within human society.
Results emphasize that AI-generated information tends to prevail in information filtering, whereas real human data is often suppressed, leading to a loss of information diversity. This trend limits next-generation model performance owing to fresh data scarcity and threatens the human information ecosystem.
Some of the limitations of our current work are mentioned in Appendix~\ref{Appendix:limitations}. 
In the future, we plan to analyze the impact of~\OurMODEL{} 
to understand information about text-specific entities~\citep{FGER,
FGERH}.

\bibliography{custom}
\bibliographystyle{colm2024_conference}
\clearpage
\appendix

\section{Ritual View of Communication}\label{sec:Ritual_view}

\citet{Carey2008Communication} conceptualized {\em "The Ritual View 
of Communication"} in his communications theory. 
They emphasized that communication is not just a medium for the 
transmission of information, but as a symbolic process that 
contributes to the construction and maintenance of social reality.

\citet{Carey2008Communication}’s theory posits that 
communication is integral to the representation, maintenance, 
adaptation, and sharing of a society's cultures over time. 
\eat{Sharing, participation, association, and fellowship are all 
central to his views.} In short, the Ritual view conceives 
communication as a process that enables and enacts societal 
transformation. 
\eat{The relevance of }This theory even extends to modern 
media forms such as newspapers and social media platforms in our 
modern communication age~\citep{CommunicationAge, 
edwards2016communication}. 
The emergence of the internet and social platforms (e.g., Facebook, 
Twitter etc.,) has further developed the ritualistic nature of communication. 
These advancements have facilitated the growth of global online 
communities by redefining their patterns of interaction
\citep{10.3389/fpsyg.2021.784907, LeeKim2014TwitterSocial}.

Similarly, generative-AI represents a profound transformation in 
the modes of human social communication and the ways humans 
interact with AI~\citep{fui2023generative,rane2023chatgpt}. 
We should regard AI, trained on extensive human civilization data, 
as an integral part of human societal information transmission, acknowledging its role in shaping and sharing the cultural and social implications of human 
society~\citep{papadimitriou2016future, rane2023chatgpt}.

\clearpage
\section{Prompt Templates}

\subsection{Answer Generation Prompt Template}
\label{appendix:Generation_Prompt_Template}
In this section, we present the prompt template for generating the \texttt{Originally Generated Answer},
the \texttt{Best Quality Answer} 
and the \texttt{Worst Quality Answer}. 

\begin{tcolorbox}[colback=gray!5!white,colframe=black!75!black,title=The prompt template for generating the \texttt{Originally Generated Answer}]
\label{table:prompttemplateOriginallyGeneratedAnswer}
Question:{query}+{detail} \\
Answer the question:
\end{tcolorbox}

\begin{tcolorbox}[colback=gray!5!white,colframe=black!75!black,title=The prompt template for generating the \texttt{Best Quality Answer}]
\label{table:prompt template best Answer}
Below is an instruction from an user and a candidate answer. Evaluate whether or not the answer is a good example of how Al Assistant should respond to the users instruction. score=5: It means it is a perfect answer from an Al Assistant. It has a clear focus on, being a helpful Al Assistant, where the response looks like intentionally written to address the user's question or instruction without any irrelevant sentences. The answer provides high-quality content, demonstrating extensive knowledge in the area, is very well written, logical, easy to follow. \\

Question: {query}+{detail} \\

Now give an example of an Al assistant answer with a score of 5 about the 
question:
\end{tcolorbox}
\begin{tcolorbox}[colback=gray!5!white,colframe=black!75!black,title=The prompt template for generating the \texttt{Worst Quality Answer}]
\label{table:prompt template worst Answer}
Provide an AI assistant response with a score of 1(lowest quality) based on the given instruction: Your example should demonstrate an incomplete, vague, off-topic, controversial, or exactly what the user asked for. \\

Question: {query}+{detail} \\

Now give the counter-example of an AI \\

Assistant response: 
\end{tcolorbox}

\subsection{AI-Washing Prompt Template}
\label{appendix:C}
We use prompts that have nothing to do with the content generated and instead have to do with the quality of the generation, as presented in the following table.

\begin{tcolorbox}[colback=gray!5!white,colframe=black!75!black,title=The prompt template for AI-washing]
\label{table:aiwashprompt}
\textbf{Prompt for ChatGPT:} \\
(en) Polish the following paragraph: \\
\{paragraph\} \\

\textbf{Prompt for SDXL:} \\
Positive: best quality, masterpiece, ultra detailed, 8K, UHD, Ultra Detailed \\Negative: worst quality, split picture, ignoring prompts, lowres 
\end{tcolorbox}

\clearpage 
\subsection{LLM Cross-scoring Prompt}
\label{appendix:Cross_scoring_Prompt}
We use the same prompt as in the work of \citet{li2023selfalignment}. as shown in the following.
\begin{tcolorbox}[colback=gray!5!white,colframe=black!75!black,title=The prompt template for evaluating answers]
\label{table:The prompt template for evaluating answers}
Below is an instruction from an user and a candidate answer. Evaluate whether 
or not the answer is a good example of how Al Assistant should respond to 
the users instruction \\

Please assign a score using the following 5-point scales \\

1: It means the answer is incomplete, vague, off-topic, controversial, or 
exactly what the user asked for For example, some content seems missing, 
the numberedlist dnot start from the beginning, the opening sentence 
repeats the user's question. Or the response is from another person's 
perspective with their personal experience (e.g. taken fmblog posts), or
looks like an answer from a forum. Or it contains promotional text, 
navigation text, or other irrelevant information \\

2: It means the answer addresses most of the asks from the user. It does 
not directly address the user's question. For example, it only provides a 
high-level instead of the exact solution to the user's question \\

3: It means the answer is helpful but not written by an Assistant. It 
addresses the basic asks of the user. It is complete and self-contained 
with the drawback that the response is not written from an assistant's 
perspective, but from other people's perspective. The content looks like 
an excerpt from a blog post, or web page, and provides search results. For 
example, it contains personal experience or opinion, mentions comments
section, or shares on socialmedia, etc. \\

4: It means the answer is written from an Al assistant's perspective with a 
clear focus on addressing the instruction. It provides a the complete, clear, 
and comprehensive response to user's question or instruction without missing 
or irrelevant information. It is well organized self-contained, and written
in a helpful tone. It has minor room for improvement, more concise and 
focused. \\

5: It means it is a perfect answer from an Al Assistant. It has a clear focus 
on, being a helpful Al Assistant, where the response looks like intentionally 
written to address the user's question or instruction without any irrelevant
sentences. The answer provides high-quality content, demonstrating extensive
knowledge in the area, is very well written, logical, easy to follow,
engaginIt means it is a perfect answer from an Al Assistant. It has a clear 
focus on, being a helpful Al Assistant, where the response looks like 
intentionally written to address the user's question or instruction without 
any irrelevant sentences. The answer provides high-quality
content, demonstrating extensive knowledge in the area, is very well written, 
logical, easy to follow, engaging, and insightful please first provide brief
reasoning you used to derive the rating score, and then write "Score: 
|rating" in the last line. \\

generated instruction: \{question\}+\{detail\} \\

answer: \{answer\}
\end{tcolorbox}

\subsection{Scoring Criteria for Human Evaluation}
Based on the modifications to the previous scoring prompts for the LLMs, we created scoring criteria for our crowdsourced annotators, as demonstrated in the following. 
\label{appendix:Scoring Criteria for Human Evaluation}
\begin{tcolorbox}[colback=gray!5!white,colframe=black!75!black,title=Scoring criteria for crowd-sourced annotators]
\label{table:Scoring criteria for crowdsourced annotations}
You are to evaluate the quality of a response given to a specific question. 
Your evaluation should consider how well the response addresses the query, 
its completeness, clarity, and relevance. \\

Scoring Scale:\\

Score 1: The response is unsatisfactory. It is incomplete, vague, unrelated 
to the question, or may simply echo the question without providing an answer. 
The content may be off-topic, contain promotional material, or resemble a 
personal opinion rather than a factual answer. \\

Score 2: The response generally relates to the question but does not directly
answer it. It may provide an overview rather than the specific details or 
solution that the question warrants. \\

Score 3: The response is useful and addresses the basic query. However, it 
may not be from the expected perspective, potentially reading like a generic 
excerpt from a blog or an article rather than a targeted answer. \\

Score 4: The response is on target, addressing the question directly and 
completely with a clear and organized presentation. Minor improvements could 
be made to enhance focus or conciseness. \\

Score 5: The response is exemplary, directly and comprehensively addressing 
the question with high-quality content. It demonstrates extensive knowledge, 
is logically structured, easy to understand, engaging, and provides insight.
Procedure for Evaluation: \\

Read the question and the corresponding response carefully.
Evaluate the response based on the above criteria. \\

Question: {query}+{detail} \\

Response: {answer} \\

Record your score : 

\end{tcolorbox}


\section{Experimental Details}

\subsection{Dataset}
\label{appendix:Dataset}
For evaluation, we manually curated text and image data sets. 
Details about these datasets are provided as follows.

\noindent {\bf QA-pairs.}
For this, we initially handpicked 100 diverse question-answer 
pairs from Stack-Overflow and Quora as the seeds. Subsequently, 
for each question in these pairs, we used large models to 
generate initial responses with instruction in Section~\ref{appendix:Generation_Prompt_Template}.
We manually screened the most answered questions in 
Stack-Overflow and Quora, including psychology, books, 
mathematics, physics, and other fields.  At the same time, 
we selected fragments from the novel corpus for anonymization 
processing to study the behavior of the language model when 
delivering real human-generated data.

\eat{
Our dataset (QA pairs) was generated by  manually processing 
selected seed data through a large language model. The seeds 
were sourced from Stack-Overflow and Quora, featuring the most 
popular questions and top-supported answers.}

These responses were further processed to curate datasets 
rated from 1 (lowest) or 5 (highest) in terms of quality,
similar to the self-alignment approach proposed by
~\citet{li2023selfalignment}.  
The prompts used for generating these diverse responses 
are detailed in Appendix \ref{appendix:Generation_Prompt_Template}. 
Final data set encompasses approximately 1,900 question-answer pairs. 
Table~\ref{tab:Appendix-data} illustrates the distribution of the data.
Formally, the dataset consists of a series of 22 tuples, each 
structured as follows:
\begin{equation}
\label{Eq:data}
    T_j = \{d, Q, D, A\} \cup \bigcup_{i=0}^{5} \{A_{\text{m}_i}, A_{\text{m}_i\text{s}_5}, A_{\text{m}_i\text{s}_1}\}
\end{equation}

\begin{table}[ht]
\centering
\resizebox{0.5\linewidth}{!}{
\begin{tabular}{l|c}
\toprule
\textbf{Data Category} & \textbf{Percentage} \\ 
\midrule
Stackoverflow QA & 30\% \\ 
Quora QA - Books & 10\% \\ 
Quora QA - Psychology & 10\% \\ 
Quora QA - Life & 10\% \\ 
Quora QA - Happiness & 10\% \\ 
Quora QA - Personal Experiences & 10\% \\ 
Quora QA - Mathematics & 10\% \\ 
\bottomrule
\end{tabular}}
\caption{Categories for QA-pairs}
\label{tab:Appendix-data}
\end{table}

\noindent {\bf Book3.}
For AI-washing experiment in Section~\ref{sec:AI_Washing}, 
the raw text dataset construction process begins with the selection 
of passages from classic literature known for their rich stylistic 
features and thematic significance, where the English dataset is 
excerpted from the pile books3~\citep{pile}, and the Chinese passages 
are selected from WebNovel. A meticulous anonymization process is 
employed to prevent the large language model from identifying the 
textual sources. This involves the alteration of recognizable names, 
places, and events.

\noindent {\bf Image-ax.}
The image dataset was constructed by \eat{carefully }selecting a 
subset of images from the \eat{comprehensive }ILSVRC data~\cite{ILSVRC15}
and web resources.
\eat{as well as other web image data, with selected} 
We select categories covering a wide range of topics and scenarios, 
in order to cover a broad range of visual features and complexity.
On the visual dataset, we sampled and cleaned the ILSVRC \citep{ILSVRC15} to ensure the diversity of image clarity and classification.

\eat{The dataset encompasses a series of 22 tuples based 
on the seed data, each structured as follows:
\begin{equation}
\begin{aligned}
    T_j = \{d, Q, D, A\} \cup \bigcup_{i=0}^{5} \{A_{\text{model}_i}, \\ A_{\text{model}_i\text{score5}}, A_{\text{model}_i\text{score1}}\}
\end{aligned}
\end{equation}}

\subsection{Computational details}
\label{Appendix:datavec-sim}
\noindent {\bf The density of cosine similarity scores} between two 
vectors $A$ and $B$ is calculated as:
\begin{equation}
\text{Cosine Similarity} = \frac{A \cdot B}{\|A\| \|B\|}
\end{equation}
where $A$ and $B$ are the embedding vectors of two paragraphs.

The density of cosine similarity scores is estimated using Kernel Density Estimation (KDE), which is given by:

\begin{equation}
\text{KDE}(x) = \frac{1}{n}\sum_{i=1}^{n} K_h(x - x_i)
\end{equation}

where,
\begin{itemize}
    \item $K_h$ is the kernel function with bandwidth $h$
    \item $x$ represents the value at which the density is estimated
    \item $x_i$ are the data points (cosine similarity scores in this case)
    \item $n$ is the number of data points.
\end{itemize}

The KDE process smoothens the discrete data points to create a continuous density curve, represented on the y-axis of Figure~\ref{fig:average_simchange}.

\noindent {\bf The average cosine similarity difference }  between two successive iteration is calculated as follows:

\begin{equation}
\Delta S = \bar{S}_i - \bar{S}_{i-1}
\end{equation}

Where:
\begin{itemize}
    \item $\Delta S$ is the average cosine similarity difference between the current text and the previous text.
    \item $\bar{S}_i$ is the average cosine similarity for current text.
    \item $\bar{S}_{i-1}$ is the average cosine similarity for previous text.
    \item For the first file comparison, $\bar{S}_{i-1}$ is assumed to be 1.
\end{itemize}

This calculation method provides a metric for assessing the change in similarity across sequential data sets, reflecting the evolution or consistency of the data characteristics.

\subsection{Exam Scenario Simulation}
\begin{figure}[ht]
\centering
\includegraphics[width=0.7\textwidth]{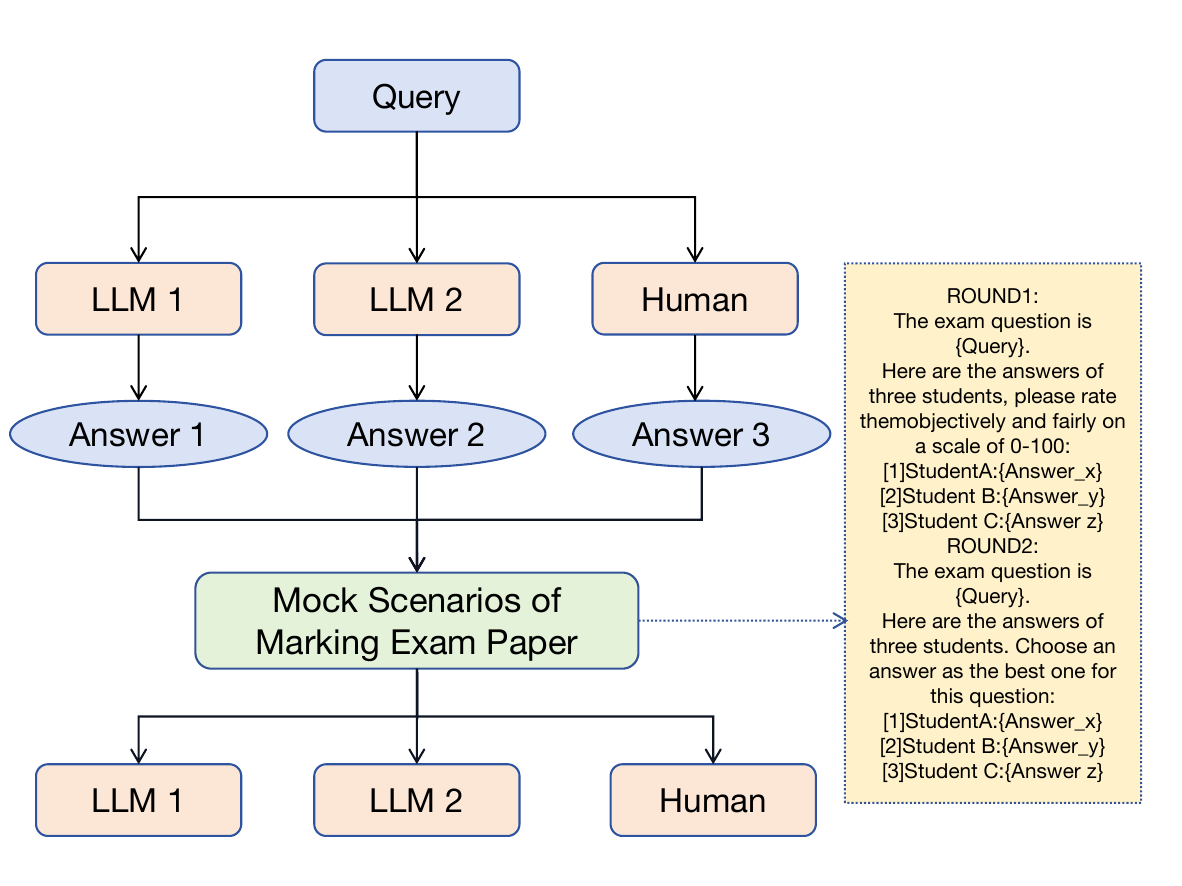}
\caption{Exam Scenario Simulation}
\label{fig:Exam_Scenario_Simulation}
\end{figure}
Figure~\ref{fig:Exam_Scenario_Simulation} displays the flowchart and prompt template for the Exam Scenario Simulation experiment.
\section{Examples of AI-washing Experiments}

\begin{table*}[t]
    \centering
    \begin{small}
    \scriptsize
    \resizebox{0.95\linewidth}{!}{
    \begin{tabularx}{13cm}{p{4cm}p{4cm}p{4cm}}
    \midrule
        \textbf{Initial Paragraph} & \bm{$N=1$} & \bm{$N=2$} \\ 
        \midrule
        An individual who was standing beyond the railing, in the free space around the marble table, and whom no one had yet caught sight of, since his long, thin body was completely sheltered from every visual ray by the diameter of the pillar against which he was leaning; this individual, we say, tall, gaunt, pallid, blond, still young, although already wrinkled about the brow and cheeks, with brilliant eyes and a smiling mouth, clad in garments of black serge, worn and shining with age, approached the marble table, and made a sign to the poor sufferer. & \hl{In the open expanse around the marble table, a figure stood just beyond the railing, elusive to the gaze of those present.} His \hl{lengthy, slender frame} remained entirely shielded from view by the pillar's breadth against which he leaned. This person, tall and lean, possessed a \hl{pallid countenance}, complemented by \hl{blond locks}. Though still in the \hl{bloom of youth}, the \hl{furrows etched across his brow and cheeks} spoke of time's passage. His eyes sparkled brilliantly, and a \hl{gentle smile graced his lips}. Draped in \hl{well-worn black serge, which bore a subdued radiance} earned through years of use, he approached the marble table and \hl{gestured to the afflicted soul}.& In the \hl{expansive openness} surrounding the marble table, a figure stood just beyond the railing, elusive to the \hl{gazes} of those present. His \hl{lengthy and slender} frame remained \hl{completely} shielded from view by the \hl{width} of the pillar against which he leaned. This \hl{tall and lean individual} possessed a pallid countenance, \hl{enhanced} by his blond locks. Although still in the \hl{prime} of youth, the \hl{lines etched} across his \hl{forehead} and cheeks spoke of the \hl{passage} of time. His eyes \hl{shimmered} with brilliance, and a gentle smile graced his lips. \hl{Adorned} in a well-worn black serge, which \hl{exuded a muted radiance} earned through years of use, he approached the marble table and \hl{gestured towards} the \hl{troubled} soul.\\ \midrule
        \bm{$N=3$} & \bm{$N=4$} & \bm{$N=5$} \\ \midrule
        In the expansive area encircling the marble table, a figure stood just beyond the railing, \hl{evading} the gaze of those present. His \hl{lithe, elongated form} remained \hl{entirely} veiled from view by the \hl{pillar’s substantial girth} against which he leaned. This \hl{individual, statuesque and slender}, possessed a \hl{pallor} that harmonized with his blond locks. Despite \hl{being} in the prime of his youth, the lines etched across his forehead and cheeks \hl{whispered} of the passage of time. His eyes \hl{gleamed} with a \hl{brilliant intensity}, and a \hl{tender} smile graced his lips. Cloaked in \hl{weathered} black serge, which \hl{bore} a \hl{subdued} luminance earned through years of wear, he approached the marble table and \hl{extended a gesture to} the afflicted soul. & In the expansive \hl{expanse} surrounding the marble table, a figure stood just beyond the railing, \hl{eluding} the gaze of those present. His lithe, elongated \hl{silhouette} remained entirely \hl{shrouded} from view by the pillar’s substantial girth against which he leaned. This individual, statuesque and slender, possessed a pallor that \hl{blended seamlessly} with his blond locks. Despite being in the \hl{zenith} of his youth, the lines etched across his forehead and cheeks \hl{murmured} of the passage of time. His eyes \hl{shone} with a brilliant intensity, and a tender smile graced his lips. Draped in weathered black serge, which bore a subdued \hl{radiance} earned through years of wear, he approached the marble table and extended a gesture to the afflicted soul.
         & In the expansive expanse surrounding the marble
        table, a figure stood just beyond the railing, \hl{eluding}
        the gaze of those in attendance. His \hl{graceful},
        elongated silhouette remained entirely shrouded
        by the substantial girth of the pillar against which
        he leaned. This person, statuesque and slender,
        possessed a \hl{complexion that seamlessly} blended
        with his blond locks. Despite being in the zenith
        of his youth, the lines etched across his forehead
        and cheeks whispered of the passage of time. His
        eyes shimmered with a brilliant intensity, and a
        tender smile graced his lips. Draped in weathered
        black serge, which bore a subdued radiance earned
        through years of wear, he approached the marble
        table and extended a \hl{reassuring} gesture to the \hl{troubled}
        soul.
 \\ \bottomrule
    \end{tabularx}}
    \end{small}
    \caption{Example of AI-Washing experiment for text from ChatGPT. $N$ represents the number of times the large language model is used for refinement, with each changed part highlighted.}
    \vspace{-3.7ex}
    \label{table:aiwashtext}
\end{table*}

\subsection{AI-washing for text data}
\label{Appendix:ai-washtext}
We report an example illustration for AI-washing for text data in Table~\ref{table:aiwashtext}.

\subsection{AI-washing for image data}
\label{Appendix:ai-washimage}

\begin{figure*}[ht]
\centering
\includegraphics[width=0.9\textwidth]{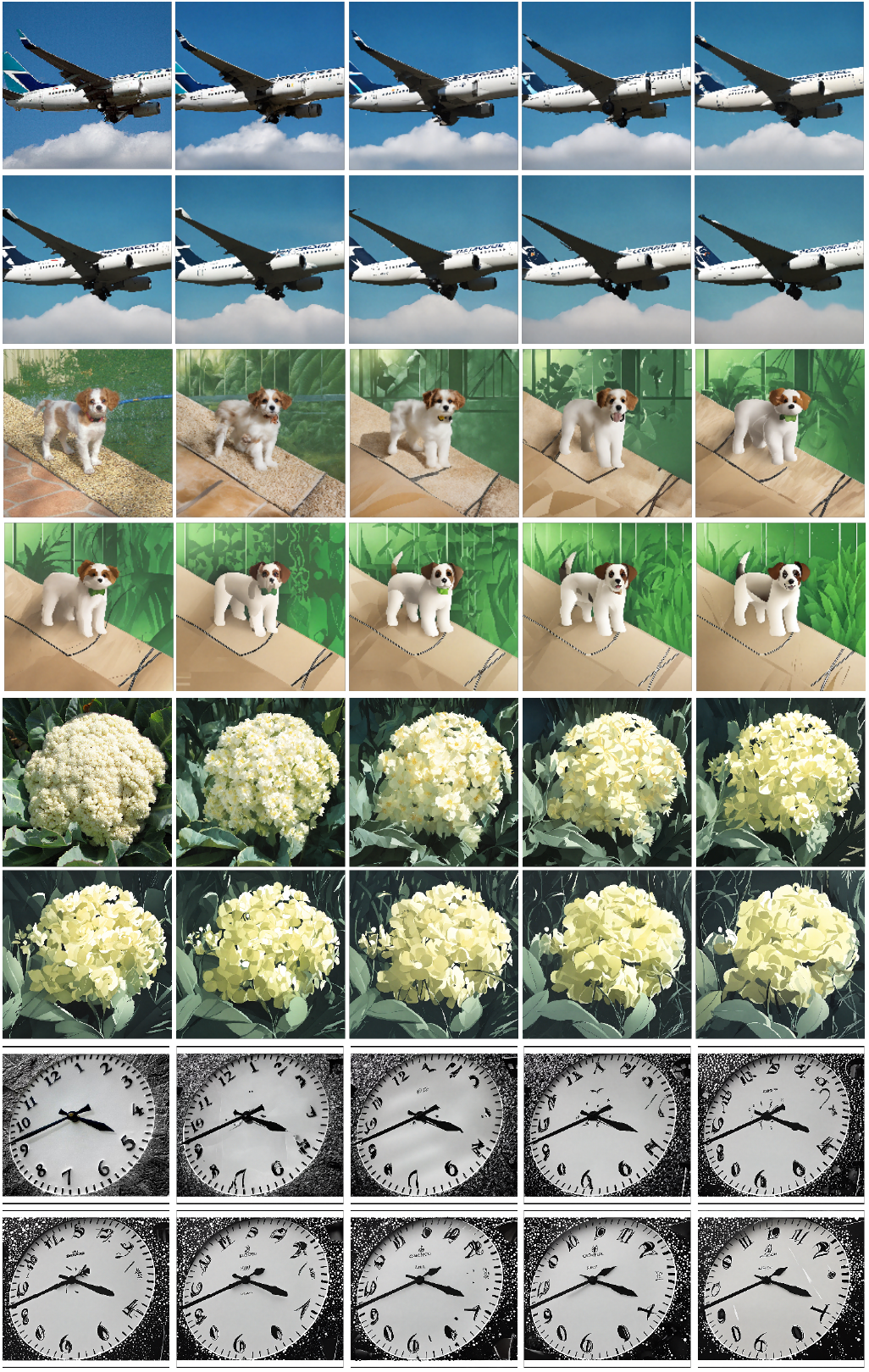}
\caption{Examples of image AI-washing experiments (part1)}
\label{fig:sdxlexample1}
\end{figure*}

\begin{figure*}[ht]
\centering
\includegraphics[width=0.9\textwidth]{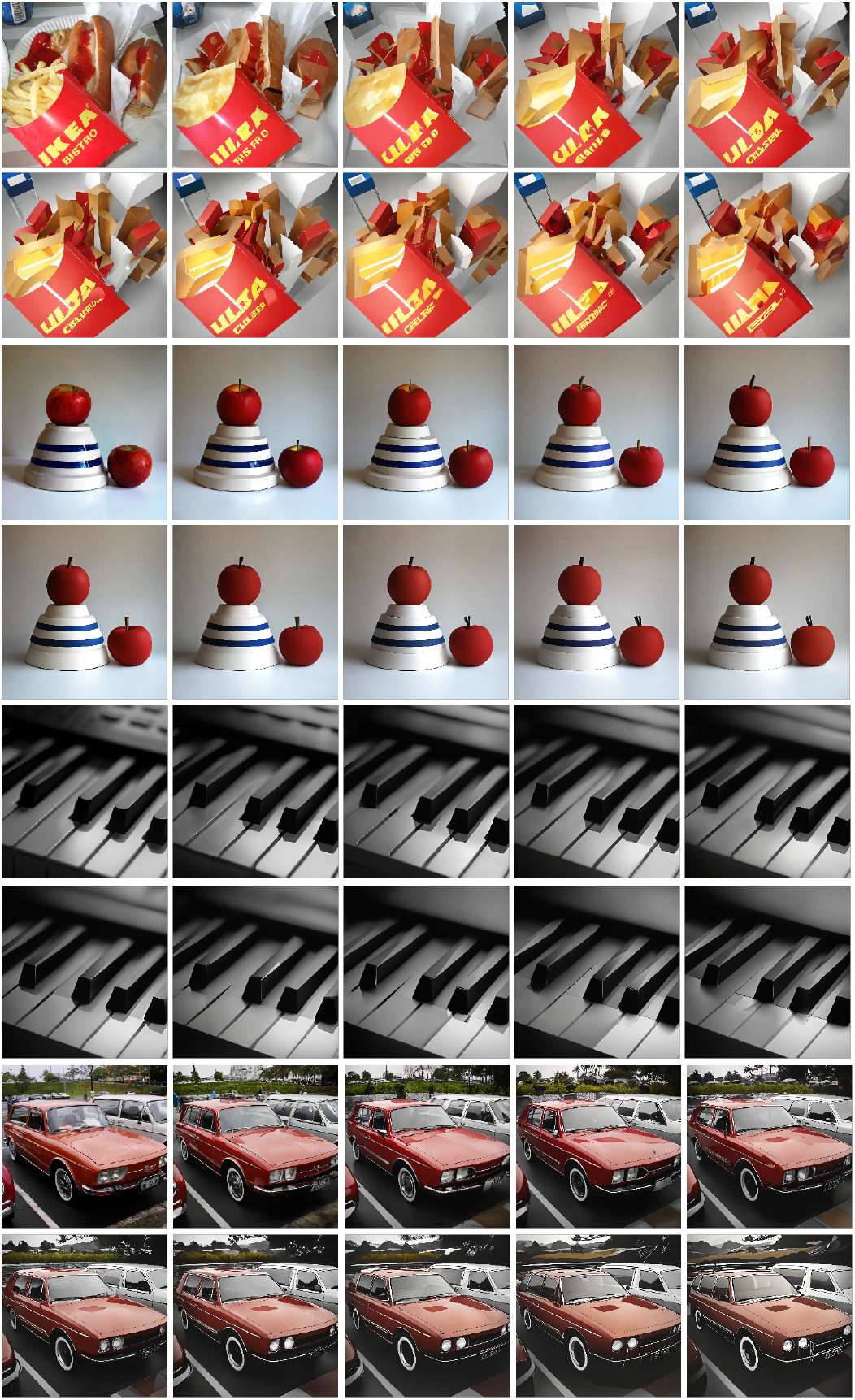}
\caption{Examples of image AI-washing experiments (part2) }
\label{fig:sdxlexample2}
\end{figure*}

We give more examples of the image AI-washing experiments in Figure~\ref{fig:sdxlexample1} and Figure~\ref{fig:sdxlexample2}, where we can observe that after iterative processing the textual parts of the images are frequently changed and fragmented, e.g., the text on the airplane, the numbers on the clock, and the letters on the potato chip packet are changed several times. The pet dog is gradually stylized as a cartoon and becomes black and white, and the cauliflower is transformed by the model into a bouquet of flowers after the first processing and is gradually stylized as a cartoon. At the same time, the model adds features to the initial image based on stereotypes from the training data, such as the logo of a clock and the logo of a car. In contrast, the overall structure, colors, and borders of the image of an apple are not significantly changed. It can be seen that the model will be affected by the model's own structure and training process when processing image features and has different enhancement or inhibition effects on different features.

\clearpage

\section{Limitations}
\label{Appendix:limitations}
\noindent {\bf Selection of Models.} 
While we have experimented with LLMs that are available, many outstanding models are worth exploring in the future. These include the GLM family of models\citep{du-etal-2022-glm}, which are known for their innovative architectures, and the MoE-structured Mixtral 8x7B\footnote{\url{https://mistral.ai/news/mixtral-of-experts/}}, etc. In addition, some open-source multilingual models are also worth investigating, such as the Qwen series of models trained on a large Chinese corpus\citep{qwen} and the Arabic model Jais\footnote{\url{https://inceptioniai.org/jais/}}. Models with different languages, parameter sizes, and architectures exhibit different behaviors. In the field of visual models, more open-source and commercial models, such as Midjourney and DALL-E 3, are worth investigating. In future research, we aim to deeply analyze the roles and characteristics of these models as an important part of human social information transfer.

\noindent {\bf Reliability of crowd-sourced Annotators.} 
A significant portion of our conclusions is derived from crowd-sourced annotators sponsored by a start-up company's data annotation department. Of these annotators, 64 \% hold graduate degrees in science and engineering, and all possess proficient bilingual reading skills in Chinese and English. However, ensuring that their existing AI knowledge does not bias their judgments remains challenging. Additionally, the distribution of our annotators in the real world varies from the general user base of generative models. There is also an ongoing debate about the reliability of crowd-sourced workers\citep{9660043,TARASOV20146190}.  \citet{veselovsky2023artificial} have discussed the behavior of annotators using LLMs for labeling, which could compromise the reliability of the results.

\noindent{\bf Privacy Issue of Large Language Models.}
Privacy concerns pose a significant limitation to the widespread adoption of Large Language Models (LLMs), despite their acknowledged capacity for knowledge comprehension \citep{yang2024moral}, and their applicability across diverse domains. Alongside privacy \citep{hu2023differentially}, issues of explainability \citep{hu2023seat,lai2023faithful} are also paramount in LLM deployment. Given that LLM applications often entail handling sensitive data \citep{xu2023llm}, effective measures are imperative to safeguard privacy. One promising avenue is the development of Differentially Private (DP) algorithms \citep{dwork2006calibrating}, which offer robust protection against identification and mitigate the risks associated with auxiliary information. Although extensive research has been conducted on DP in machine learning \citep{hu2022high,wang2020differentially,wang2023generalized,su2022faster,hu2023privacy} and deep learning \citep{xiang2024does,xiang2023practical,shen2023differentially}, the focus has predominantly been on tabular or image data. Unfortunately, there is a noticeable dearth of attention towards adapting DP algorithms to the Natural Language Processing (NLP) and text data. Given the unique challenges posed by textual information, specialized privacy-preserving techniques tailored to NLP tasks are essential. Addressing this gap is pivotal for enhancing the privacy protections of LLMs and ensuring their ethical deployment across various domains. However, further exploration of this area is warranted and will be left for future research endeavors.


\eat{The detailed explanation about the mathematical notation 
in Equation~\ref{Eq:data} is provided in Appendix~\ref{appendix:Dataset}
Appendix~\ref{appendix:Dataset} provides explanations of the 
corresponding mathematical notation, showing more details about 
the distribution of the dataset.}

\eat{where \( j \) indexes the tuple within the dataset. Each element of the tuple is defined as:
\begin{itemize}
    \item \( d \): The domain of the question-answer pair which provides context for the question classification.
    \item \( Q \): The question posed by a user that serves as a direct input for model-generated answers.
    \item \( D \): Document-related information that provides background knowledge necessary for answering \( Q \).
    \item \( A \): The human answer that received the most endorsements for question \( Q \), serving as a benchmark for answer quality.
\end{itemize}
For each language model \( \text{model}_i \), where \( i \) ranges from 0 to 5, representing one of six different large language models:
\begin{itemize}
    \item \( A_{\text{model}_i} \): The initial answer generated by model \( i \).
    \item \( A_{\text{model}_i\text{score5}} \): The highest quality answer generated by model \( i \), according to prompts.
    \item \( A_{\text{model}_i\text{score1}} \): The lowest quality answer generated by model \( i \), according to prompts.
\end{itemize}}
\end{document}